\documentclass[12pt]{article}
\usepackage{amsmath,amssymb,fullpage,graphicx,amscd,array,amsfonts,mathtools,amsthm}
\usepackage{dsfont}
\usepackage{xcolor}
\usepackage{url}
\usepackage[english]{babel}
\usepackage{graphicx}
\usepackage{longtable}
\usepackage{caption}
\usepackage{appendix}
\usepackage[font=footnotesize]{caption}
\def\pf#1{\vspace{1mm}\noindent{\it #1}\quad}
\def\leq{\leqslant}
\def\geq{\geqslant}

\RequirePackage{geometry}
\allowdisplaybreaks
\newtheorem{theorem}{Theorem}[section]

\newtheorem{example}{Example}[section]

\newtheorem{definition}{Definition}[section]

\begin{document}

%
%
%
%
%
\renewcommand{\thefootnote}{\alph{footnote}}

\begin{center}
{\LARGE The Spaces of Data, Information, and Knowledge}\\

\vspace{5mm}

Xiaoyu Chen\footnote{\it SKLSDE - School of Computer Science and Engineering, Beihang University, Beijing 100191, China. E-mail: franknewchen@gmail.com} and Dongming Wang\footnote{\it LMIB - School of Mathematics and Systems Science, Beihang University, Beijing 100191, China}$^,$\footnote{\it Centre National de la Recherche Scientifique, 3 rue Michel-Ange, 75794 Paris cedex 16, France}

\vspace{3mm}

\end{center}

\begin{abstract}
\centering
\vspace{3 true pt}
\begin{minipage}{0.8\textwidth}
\indent
We study the data space $D$ of any given data set $X$ and explain
how functions and relations are defined over $D$. From $D$ and for a
specific domain $\Delta$ we construct the information space $I$ of
$X$ by interpreting variables, functions, and explicit relations
over $D$ in $\Delta$ and by including other relations that $D$
implies under the interpretation in $\Delta$. Then from $I$ we build
up the knowledge space $K$ of $X$ as the product of two spaces $K_T$
and $K_P$, where $K_T$ is obtained from $I$ by using the induction
principle to generalize propositional relations to quantified
relations, the deduction principle to generate new relations, and
standard mechanisms to validate relations and $K_P$ is the space of
specifications of methods with operational instructions which are
valid in $K_T$. Through our construction of the three topological
spaces the following key observation is made clear: the retrieval of
information from the given data set for $\Delta$ consists
essentially in mining domain objects and relations, and the
discovery of knowledge from the retrieved information consists
essentially in applying the induction and deduction principles to
generate propositions, synthesizing and modeling the information to
generate specifications of methods with operational instructions,
and validating the propositions and specifications. Based on this
observation, efficient approaches may be designed to discover
profound knowledge automatically from simple data, as demonstrated
by the result of our study in the case of geometry.

\medskip

\textbf{Keywords} data space, declarative knowledge, deduction principle, implied
knowledge, induction principle, information retrieval, knowledge
discovery, procedural knowledge, quantified relation
\end{minipage}
\end{abstract}

\renewcommand{\thefootnote}{\arabic{footnote}}
\setcounter{footnote}{0}

\section{Introduction}

With increasingly wide use of digital devices and networks, more and
more scientific explorations and social activities are carried out
electronically, where the involved objects, phenomena, and behaviors
of interest are measured and presented in the form of data. Thus
analyzing and modeling data, retrieving information from data, and
discovering knowledge that data imply become essential tasks. In
fact, data mining, machine learning, information retrieval, and
knowledge discovery are some of the directions of research and
development which have been given high priority in various national
programs~\cite{sm}. Let us address three issues in more
detail.

\medskip\noindent
(1) Data Management. Data are acquired, collected, and recorded in
digital databases with respect to different aspects of concern, such
as time, location, state, and relation, of observed objects. Data
models and schemas need be well designed in order to provide
manipulable structures for efficiently storing, retrieving,
exchanging, and acting on data. Visualization techniques should be
capable of representing the underling features and properties of
data in the form of intuitive diagrams with dynamic animation. Data
management techniques should be available as fundamental utilities
in dealing with the activities of observation, measurement, and
experiment. It becomes more and more difficult to process
collections of complex data with large volume, high velocity of
change, uncertain veracity, and variety of types, called big data
\cite{bigdata}, using traditional methodologies and existing
database management tools. There is a need of substantially new
ideas, techniques, approaches, and systems for the management of big
data.

\medskip\noindent
(2) Information Retrieval. Information is implied in data and may be
retrieved manually or mechanically through analyzing, modeling,
computing, and learning. Valuable information depicts the essence of
observed phenomena and can thus be used to predict the changing of
the phenomena and to study the properties of the objects under
observation. A large variety of methods, techniques, and tools have
been developed in the fields of data mining, knowledge discovery,
pattern recognition, and machine learning for retrieving information
from data (see, e.g., the top 10 influential data mining algorithms
identified by the research community~\cite{DMtop10} and ensemble
methods of effectively combining multiple algorithms and
techniques~\cite{Zhou12}), with applications to information
dissemination and the analysis of crowd behaviors and social
networks~\cite{sn1,sn2,Han}. Current investigations tend to be more
expertise-oriented and more problem-centered, as witnessed by the
publications of a series of conference proceedings and journals. An
integrated study of qualitative and quantitative methods from
natural sciences with cognitive science, psychology, and human
behaviors is likely to help enhance our understanding of information
retrieval~\cite{DMsurvey}, hopefully resulting in revolutionary
approaches for intelligent retrieval of information from big data.

\medskip\noindent
(3) Logical Reasoning. The discovery of scientific knowledge is
inseparable from logical reasoning. The principle of induction
points out how to formulate conjectures as to acquire knowledge
based on experience of a few existing cases~\cite{induction}.
On the other hand, the inductive logic provides a less-than-certain
inference mechanism of evidential support by using
probability~\cite{induction1}. Holding a controversial view on the
function of induction, Popper~\cite{Karl} argued that scientific
theories are not inductively inferred from experience, nor is
scientific experimentation carried out with a view to verifying or
finally establishing the truth of theories; rather, all knowledge is
provisional, conjectural, hypothetical, disprovable rather than
provable. In his view, scientific discovery is a deductive process
in which scientists formulate hypotheses and theories that they test
by deriving particular observable consequences, modify falsified
theories based on empirical facts, and create new theories that
corroborate the necessary predictions. Along this line of thought,
Li \cite{Li} developed an operable revision calculus to deal with
refutations and a logical framework for formalizing the process of
scientific discovery.

\medskip
Knowledge may be discovered from data via information retrieval. To
understand the process of knowledge discovery, one needs to study
the properties of data, information, and knowledge and to clarify
their relationship. Such studies are also necessary for the
establishment of theoretical foundations for the sciences of data
and knowledge. This paper presents the result of our initial study
on what we call \emph{implied knowledge discovery} by constructing
three topological spaces for data, information, and knowledge. A key
idea that is responsible for the richness of the constructed space
of knowledge is to generate data-implied knowledge by applying
meta-knowledge (notably the induction and deduction principles) and
domain knowledge. We provide formal definitions for various concepts
and theorems to describe features and relations of the three spaces
and objects therein. According to the process of construction of the
three spaces, efficient approaches may be designed to discover
profound knowledge automatically from simple data. The interested reader is encouraged to consult~\cite{csw14} for one such approach, which is capable of generating nontrivial geometric theorems from images of diagrams.

\section{The Space of Data}

We start by recalling a few standard concepts from the area of data
analysis.

\begin{definition}[Data point]\rm
A data point is a set of one or more measurements on a single member
of a set of observed objects.
\end{definition}

\begin{definition}[Data set]\rm
A data set is a collection of data points.
\end{definition}

Let $X$ be any given finite nonempty data set.

\begin{definition}[Data space]\rm
The data space $D$ of $X$ is the set $X$ of points endowed with a
family $\tau$ of subsets of $X$ such that (1) both $\emptyset$ and
$X$ are elements of $\tau$, (2) any union of elements of $\tau$ is
an element of $\tau$, and (3) any intersection of finitely many
elements of $\tau$ is an element of $\tau$. Each element in $\tau$
is called an open set and $\tau$ is called the data structure of
$D$.
\end{definition}

\begin{theorem}
 The space $D$ defined above is a finite topological space.
\end{theorem}

\pf{Proof.} By definition of topology~\cite{topology}, it can be easily proved that $\tau$ is a topology on $X$. Thus, $D$ is a
topological space for which there are only finitely many data points.

Let $n$ be a positive integer and $T\subset \tau^n$.

\begin{definition}[Data Function]\rm
A data function is a map
\[\begin{array}{rl}
f: & T \mapsto \tau \\
& (t_1,\ldots,t_n) \rightarrow f(t_1,\ldots,t_n).
\end{array}\]
\end{definition}

\begin{definition}[Data Relation]\rm
A data relation is a map
\[\begin{array}{rl}
R: & T \mapsto \{\textrm{true},\textrm{false}\} \\
& (t_1,\ldots,t_n) \rightarrow R[t_1,\ldots,t_n].
\end{array}\] For $n=2$, the relation $R[t_1,t_2]$ is sometimes written as $t_1R\,t_2$.
\end{definition}

\begin{example}\rm
Given
$t,s\in \tau$, the data function \texttt{intersection} maps $(t,s)$ to \texttt{intersection}$(t,s)$, which is written usually as $t \cap s$. There are two kinds of data relations: propositional (without using
variables, e.g., $\{3\}\subset\{3,7\}$) and quantified (to
represent a set of propositional relations, e.g., $\forall t\in \tau\, \exists s \in \tau\,(t\cup s = X)$).
\end{example}

For any fixed data set $X$, one can define different data spaces with different topologies. One can also define different binary relations (called preorders) on $X$ that are reflexive and transitive.
For any preorder $\preceq$ on $X$, there is a data space with topology $\tau$ such that every upper set $U$ of $X$ with respect to $\preceq$ (i.e., if $x\in U$ and $x\preceq y$ then $y\in U$) is an open set of $\tau$. Conversely, for any data space with topology $\tau$, there is a preorder $\preceq$ on $X$ such that $x\preceq y$ if $x$ is in the closure of $\{y\}$ in $\tau$. Therefore, the binary relations are in one-to-one correspondence with the data spaces.

In what follows, we list some properties of data spaces which are useful for data
analysis.
\begin{itemize}
\item[(1)] Compactness. Every data space is compact (because each open cover of it has a finite subcover).
\item[(2)] Separability. For any data space $D$ of $X$, if $D$ is $T_1$
(i.e., for every pair of distinct data points in $D$, each point has
an open neighborhood not containing the other), then $D$ must be
discrete (i.e., the data structure of $D$ is the power set of $X$).
\item[(3)] Connectivity. A data space is connected if and only if the associated graph with respect to its corresponding preorder is path-connected.
\item[(4)] Metrizablity. A data space is metrizable if and only if it is discrete.
\end{itemize}

If a data space of $X$ is metrizable, then one can introduce a metric on $X$
as follows.
\begin{definition}[Metric]\rm
A metric on $X$ is a map $d: X^2 \mapsto \mathds{R}$
(where $\mathds{R}$ denotes the field of real numbers) which
satisfies the following conditions: for all $x$, $y$, $z\in X$,
(1)~$d(x, y)\geq 0$; (2)~$d(x, y) = 0$ if and only if $x = y$;
(3)~$d(x, y) = d(y, x)$; (4)~$d(x, z) \leq d(x, y) + d(y, z)$.
\end{definition}

\begin{definition}[Similarity]\rm
Two data points $x$ and $y$ in $X$ are said to be $\epsilon$-similar, denoted as $x\sim_\epsilon y$, if $d(x,y) \leq \epsilon$, where $d$ is a metric\footnote{When the metric is Euclidean distance, $\epsilon$-similarity measures the closeness of data points in $X$.} on $X$ and $\epsilon$ is a given threshold. If $x$ and $y$ are not $\epsilon$-similar, then we write $x\not\sim_\epsilon y$.
\end{definition}

\begin{definition}[Cluster]\rm
A subset $C$ of $X$ is called a cluster of $X$ if
\begin{enumerate}
 \item[(a)] for any two data points $x_0,x\in C$, there exist $x_1,\ldots,x_n\in C$ such that $x_{i}\sim_\epsilon x_{i-1}$ for $i=1,\ldots,n$ and $x\sim_\epsilon x_n$;
 \item[(b)] for any $x\in C$ and $y\in X\setminus C$, $x\not\sim_\epsilon y$,
\end{enumerate}
where $\epsilon$ is a given threshold.
\end{definition}

Obviously, any two clusters of $X$ are disjoint. It is also easy to prove the following.

\begin{theorem} For any given finite data set $X$ and threshold $\epsilon$, there are finitely many clusters $C_1,\ldots,C_m$ such that $X = C_1\cup \cdots \cup C_m$.
\end{theorem}

When $X$ is a subset of the Euclidean space and $d$ is the Euclidean distance, $X$ can be converted into a graph by taking the points in $X$ as its vertices and connections of proximate vertices as its edges. The obtained graph can be turned into a simplicial complex of $X$ by gluing together simplices. The constructed simplicial complex is a topological space. Therefore, methods from algebraic topology can be used to study the simplicial complex of $X$ (see~\cite{Carl} for more details).

The data structure of a data space, in which useful domain
information is implied, plays an important role in depicting the
features of the space.

\section{Domain and Interpretation}

Let $\Delta$ be an arbitrary but fixed domain.

\begin{example}\label{exmp}\rm
The data set $X=\{3, 7, 11, 23\}$ has no meaning. It may be
interpreted as a set of strings of characters, or a set of
mathematical numbers, or a set of identifiers for different
athletes.
\end{example}

\begin{definition}[Domain object]\rm
A domain object is an object of study that has clear meaning in
$\Delta$.
\end{definition}

\begin{definition}[Domain function]\rm
A domain function is a function that is defined over a subdomain of
$\Delta$ and has clear meaning in $\Delta$.
\end{definition}

\begin{definition}[Domain relation]\rm
A domain relation is a relation among domain objects that has clear
meaning in $\Delta$.
\end{definition}

\begin{definition}[Propositional and quantified relation]\rm
A domain relation which does not involve any quantifier is called a
propositional relation. A quantified relation is a relation which
involves at least one of the quantifiers $\forall$ and $\exists$ to
represent a set of propositional relations.\footnote{For the sake of
convenience, we call any first-order logical formula over $\Delta$ a
domain relation; so a proposition is also a domain relation.}
\end{definition}

Domain functions and relations are introduced usually by definitions
in the domain. There are two types of domain objects: primitive
objects and derived objects. Primitive objects may be defined
informally, while derived objects are defined through domain
functions on primitive objects and already defined derived objects.
Similarly, there are two types of domain relations: primitive
relations and derived relations. The former may be defined
informally, while the latter are defined through domain functions on
primitive relations and already defined derived relations. To
facilitate the study of the domain, let the set of primitive objects
and relations be well chosen and then fixed.

\begin{definition}[Operation]\rm
Let $y=f(x_1, \ldots, x_n)$ be a function defined over a subdomain
$\delta$ of $\Delta$. For any given values
$\bar{x}_1,\ldots,\bar{x}_n\in\delta$, the evaluation
$f(\bar{x}_1,\ldots,\bar{x}_n)$ is called an operation in
$\Delta$.
\end{definition}

\begin{definition}[Instruction]\rm
An instruction is a specification about where, when, and how to
perform a sequence of operations in $\Delta$.
\end{definition}

\begin{definition}[Method]\rm
A method consists of a specification about what is given and what is
the goal to be achieved and a sequence of instructions on how to
achieve the goal step-by-step using what is given.
\end{definition}

Let $D^*$ be a set consisting of open sets of data space $D$ as well as data functions and relations on the open sets.

\begin{definition}[Interpretation]\rm
An interpretation in $\Delta$ is a map from $D^*$ to $\Delta$ that
maps each open set of $D$ to a domain object (or a set of domain objects) in
$\Delta$, each data function in $D^*$ to a domain function in $\Delta$, and each
data relation in $D^*$ to a domain relation in $\Delta$.
\end{definition}

\begin{definition}[Implied relation]\rm
Any set of domain relations in $\Delta$ which can be obtained from $D^*$ by
means of interpretation in $\Delta$ is a set of $D$-implied
relations. Furthermore, any set of domain relations in $\Delta$ which can be deduced from sets of
$D$-implied relations is also a set of $D$-implied relations.
\end{definition}

To retrieve information from $X$, one needs to mine domain objects
and $D$-implied relations.

\section{The Space of Information}

\begin{definition}[Piece of information]\rm
A piece of information in $\Delta$ is a pair $\langle \mathcal{O}, \mathcal{R}\rangle$,
where $\mathcal{O}$ is a set of domain objects in $\Delta$ and $\mathcal{R}$ is a set of domain
relations in $\Delta$ which involves the objects in $\mathcal{O}$.
\end{definition}

\begin{example}\rm
Let $X$ in Example~\ref{exmp} be interpreted as a set $X'=
\{3,7,11,23\}$ of mathematical numbers (i.e., $\Delta$ is
mathematics). Then $X'$ with the relations that (1) the average (which is
a derived object in $\Delta$) of the numbers in $X'$ is equal to
$11$ and (2) every number in $X'$ is prime is a piece of information.
\end{example}

Each piece of information can be presented in a standard mathematical
structure (such as an ordered set, a table, a tree, or a graph).

\begin{definition}[Implied information]\rm
Any piece of information in $\Delta$ which can be obtained from $D$
by means of interpretation in $\Delta$ is a piece of $D$-implied
information. Any piece of information in $\Delta$ which can be deduced from
pieces of $D$-implied information is also a piece of $D$-implied
information.
\end{definition}

Pieces of $D$-implied information may be retrieved from $D$ and thus
from the given data set $X$ interpreted in $\Delta$. To retrieve
information from $X$, the domain $\Delta$ must be specified or
detected.

\begin{example}\rm
Assume that in the domain of administration, two open sets of a data
space are interpreted into $C_d$ and $B_d$. $C_d$, together with
relations involving elements of $C_d$, is interpreted as a piece of information
$C_i$ for China and $B_d$, together with relations
involving elements of $B_d$, is interpreted as a piece of
information $B_i$ for Beijing. Then $C_i$ and $B_i$, together with
$D$-implied information involving elements of $C_d$ and/or $B_d$,
form a piece of information $A_i$ for China and Beijing.

$A_i$ may contain such relations
as ``China is a country,''
``Beijing is a city of China,'' and ``Beijing is the capital of
China.'' These relations are propositional and involve only the
constants ``China'' and ``Beijing'' (without variables).
\end{example}

\begin{definition}[Information space]\rm
The information space $I$ of $X$ for $\Delta$ is the set $S$ of pieces of $D$-implied information in $\Delta$ endowed with a
topology $\tau_I$ on $S$. Here $\tau_I$ is called the information
structure of $I$.
\end{definition}

There are different topologies, such as cofinite topology and
discrete topology, which can be defined on $S$. In particular, as
deductive relation $\rightarrow$ on $S$ ($a\rightarrow b$ means that $b$ can be deduced from $a$ where $a,b\in S$) forms a preorder, a topology $\tau_\rightarrow$
can thus be defined on $S$ with respect to $\rightarrow$.

An information space $I$ of $X$ for $\Delta$ is said to be induced from a data
space $D$ of $X$ if the information structure $\tau_{I}$ for $I$ is
constructed as follows:
\begin{itemize}
\item[(a)] if $u \in \tau_{I}$, then $u$ is a piece of $D$-implied information;
\item[(b)] if $u,v\in \tau_{I}$, then $u\cap v\in \tau_{I}$;
\item[(c)] if $u,v\in \tau_{I}$, then $u\cup v\in \tau_{I}$.
\end{itemize}
Each open set of an information space induced from a data space is a
finite family of pieces of D-implied information with respect to the
same set of objects in $\Delta$.

Information is not necessarily true. It becomes knowledge when
validated. For example, when ``Beijing is the capital of China'' is
validated, it becomes part of a knowledge object.

\section{The Space of Knowledge}

Knowledge is formulated from information spaces by means of
induction, deduction, synthesis, modeling, and validation.

\begin{definition}[Induction principle \cite{induction}]\rm
The principle of induction is a law to extrapolate from given
information (called premises) and predict things containing more
information than the premises make available. Let each
domain object in $\Delta$ be an instance of a concept or a class.
Then the principle of induction may be stated as follows.
\begin{itemize}
\item[(a)] The greater the number of pieces of information in the form of
$\langle \{o_1, o_2, \ldots, o_n\}, R \rangle$ is,
where each
$o_i$ is an instance of a class $C_i$, the more probable it is (if
no piece of information of failure of the relation $R$ is known)
that the relation $R$ holds among all the instances of $C_i$
for $1\leq i\leq n$.
\item[(b)] Under the same circumstances, a sufficient number of pieces of information
on the relation $R$ among some instances of $C_1,\ldots,C_n$ will make it nearly
certain that the relation $R$ among other instances of $C_1,\ldots,C_n$ is always
satisfied, and will make this general law approach certainty without limit.
\end{itemize}
\end{definition}

The induction principle may be used as a method to generalize
propositional relations to quantified relations and as a scheme for
the generation of induction proofs. Induction in a narrow sense
refers to inferences with less than 100\% probability because the
conclusion is tentatively valid, provided that and so long as no
cases are found that belie it; whereas deduction refers specifically
to inferences with 100\% probability~\cite{deduction}.

\begin{definition}[Deduction principle \cite{deduction}]\rm
The principle of deduction is a law asserting that new relations or
conclusions can be logically deduced from already established
premises.
\begin{itemize}
\item[(a)] The conclusion must be fully justified by the premises.
\item[(b)] The conclusion is sure and immutable, so long as no new information contradicts the premises.
\end{itemize}
\end{definition}

The deduction principle may be used as a method to derive new
information from known pieces of information and as a scheme for the
generation of new relations (propositions in $\Delta$).

\begin{definition}[Modeling principle]\rm
The principle of modeling is a law pointing out that models for
phenomena and their behaviors can be established from the
information on instances of the phenomena and their behaviors.
\begin{itemize}
\item[(a)] Initial models are formulated with embedded parameters according to the
information on the instances in analog to known models for similar
phenomena.
\item[(b)] The formulated models may be verified, modified, or
improved iteratively through optimization of the parameter values by
using additional information on the phenomena and their behaviors.
\end{itemize}
\end{definition}

The modeling principle may be used to design schemes for the
generation of algorithmic methods.

\begin{definition}[Validation]\rm
A piece of information, a proposition, or a method may be validated
by belief, by assumption, by proof, or by verification.
\end{definition}

Proof may be deterministic or probabilistic and verification may be
exhaustive or for samples in the domain. Statistical verification
may be used for statements involving vague words such as ``most,''
``almost,'' and ``very.'' Methods are validated usually by proofs or
verifications for correctness.

\begin{definition}[Knowledge object]\rm
A knowledge object is a description of a piece of information
(representing a segment of a fact or a phenomenon), or a
definition (of a function or a relation), or a proposition
(representing a general law), or a method in the given domain,
\ldots, which has been validated.
\end{definition}

There are mainly two types of knowledge objects, declarative and
procedural.

\begin{definition}[Declarative knowledge object]\rm
A knowledge object is said to be declarative if it declares a
propositional or quantified relation.
\end{definition}

\begin{definition}[Procedural knowledge object]\rm
A knowledge object is said to be procedural if it specifies the
functionality of a segment of a method in terms of input and output
and provides the sequence of instructions on how to produce the
output from the input.
\end{definition}

The record of a sequence of operations performed according to the
instructions provided in a procedural knowledge object for a
particular input may also be considered as a knowledge object. Such
knowledge objects include sequences of computations and proofs of
theorems produced by general or particular methods and are
secondary. They are also called procedural knowledge objects.

By synthesis we mean the generation of procedural knowledge objects
from declarative ones. Validated pieces of information together with
their extensions made by using induction, deduction, modeling,
synthesis, and validation form the space of knowledge.

\begin{example}\rm
Refer to the previous example and let ${\rm HasCapital}(x)$ denote
``$x$ has a capital.'' From the relations contained in $A_i$, one
can conclude that ${\rm HasCapital}({\rm China})$ by deduction and
conjecture that ``every country has a capital'' by induction. The
conjecture can be formulated as $\forall x\in G\,({\rm
HasCapital}(x))$, where $G$ denotes the set of all countries. When
the conjecture is validated, it becomes part of a knowledge object
in the domain of administration.
\end{example}

\begin{definition}[Declarative knowledge space]\rm
The declarative knowledge space $K_T$ of $X$ for $\Delta$ is the set
of declarative knowledge objects, which can be obtained from $I$ for
$\Delta$ by applying the induction principle, the deduction
principle, and validation mechanisms, endowed with a topology. The topology is called the knowledge structure of $K_T$.
\end{definition}

\begin{definition}[Procedural knowledge space]\rm
The procedural knowledge space $K_P$ of $X$ for $\Delta$ is the set
of procedural knowledge objects, which are valid in $K_T$, endowed
with a topology. The topology is called the knowledge structure of $K_P$.
\end{definition}


\begin{definition}[Derivation relation]\rm A derivation relation
$\twoheadrightarrow$ is a binary relation on $K_T$ and $K_P$. For
any two knowledge objects $o_1$ and $o_2$, if $o_2$ is obtained on the basis of $o_1$, then we say that $o_2$ is derived from $o_1$, denoted as
$o_1\twoheadrightarrow o_2$.
\end{definition}

The derivation relation induces a partial order because it
satisfies the following conditions for all $o_1$, $o_2$, and $o_3$
in $K_T$ and $K_P$: (1) $o_1\twoheadrightarrow o_1$; (2) if
$o_1\twoheadrightarrow o_2$ and $o_2\twoheadrightarrow o_1$, then
$o_1=o_2$; (3) if $o_1\twoheadrightarrow o_2$ and
$o_2\twoheadrightarrow o_3$, then $o_1\twoheadrightarrow o_3$.
Therefore, the derivation relation also induces a preorder.
Hence we can introduce knowledge structures
$\tau_{T\twoheadrightarrow}$ and $\tau_{P\twoheadrightarrow}$ to
define the declarative knowledge space $K_{T\twoheadrightarrow}$ and
the procedural knowledge space $K_{P\twoheadrightarrow}$,
respectively, because every open set of
$K_{T\twoheadrightarrow}$ and $K_{P\twoheadrightarrow}$ is an upper
set with respect to $\twoheadrightarrow$.

\begin{definition}[Knowledge space]\rm
The knowledge space $K$ of $X$ for $\Delta$ is the product of the
declarative knowledge space $K_T$ and the procedural knowledge space
$K_P$, denoted as $K_T\times K_P$.
\end{definition}

\begin{definition}[Section of knowledge]\rm
A subset $k$ of $K_{T\twoheadrightarrow}$ or $K_{P\twoheadrightarrow}$ is called a section of knowledge if
\begin{itemize}
\item[(a)] $k$ is a singleton set containing only one element $o$ from which no other element is derived
and there is also no other element from which $o$ is derived; or
\item[(b)] for any two knowledge objects $o_0,o_m\in k$, there exist
$o_1,\ldots, o_{m-1}\in k$ such that $o_{j}$ is derived from
$o_{j-1}$ for $j=1, \ldots, m$; or
\item[(c)] for any two knowledge objects $o,o_0\in k$, there exist $o_m,\ldots,
o_1\in k$ such that $o$ is derived from $o_m$ and $o_{j}$ is derived
from $o_{j-1}$ for $j=m, \ldots, 1$.
\end{itemize}
A section $k$ of knowledge is said to be complete if $k$ cannot be
enlarged. Each section of knowledge is an open set of
$K_{T\twoheadrightarrow}$ or $K_{P\twoheadrightarrow}$.
\end{definition}

\begin{theorem} If the knowledge space $K_{T\twoheadrightarrow}$ or
$K_{P\twoheadrightarrow}$ is finite, then any set of the space
may be uniquely decomposed into finitely many complete sections of
knowledge.
\end{theorem}

\pf{Proof.} Let $k$ be a set of $K_{T\twoheadrightarrow}$ or
$K_{P\twoheadrightarrow}$. If there exists an element $o\in k$ such
that no other element is derived from $o$ and there is no other
element from which $o$ is derived, then $k$ can be decomposed
into a section of knowledge $\{o\}$ and the set $k\setminus\{o\}$.
Otherwise, as the derivation relation $\twoheadrightarrow$ is a
partial order, $k$ with $\twoheadrightarrow$ can be represented as
finitely many disjoint directed acyclic graphs. Each directed
acyclic graph can be uniquely decomposed into finitely many longest
chains of which each corresponds to a complete section of knowledge.

Consider the set
\[S_K=\left\{k_T\otimes k_P\left| \begin{array}{r}
k_T\subset K_T, k_P\subset K_P, \forall a\in k_T \exists b\in
k_P\,(a\twoheadrightarrow b),\\ \forall b\in k_P\exists
a\in k_T\,(b\twoheadrightarrow a)\;\end{array}\right.\right\}\] of subspaces of $K$. An
element $k_T\otimes k_P$ of $S_K$ is called a \emph{dual section} of
knowledge if both $k_T$ and $k_P$ are sections of knowledge. A dual
section $k_T\otimes k_P\in S_k$ of knowledge is said to be complete
if neither $k_T$ nor $k_P$ can be enlarged. A complete dual section
of knowledge is also called a \emph{chapter} of knowledge. A chapter of
knowledge may be decomposed into dual sections of knowledge.

\medskip

\noindent \textbf{Assumption.} For each knowledge object $o_T$ in
$K_{T\twoheadrightarrow}$, there exists at least one knowledge
object $o_P$ in $K_{P\twoheadrightarrow}$ such that
$o_T\twoheadrightarrow o_P$ or $o_P\twoheadrightarrow o_T$; for each
knowledge object $o_P$ in $K_{P\twoheadrightarrow}$, there exists at
least one knowledge object $o_T$ in $K_{T\twoheadrightarrow}$ such
that $o_P\twoheadrightarrow o_T$ or $o_T\twoheadrightarrow o_P$.

\medskip
Under the above assumption, we can prove the following theorem.

\begin{theorem}
If the knowledge space $K_\twoheadrightarrow=K_{T\twoheadrightarrow}\times K_{P\twoheadrightarrow}$ is finite, then $K_\twoheadrightarrow$ may be decomposed into finitely many chapters of knowledge. The decomposition is unique.
\end{theorem}

\pf{Proof.} As the knowledge space $K_\twoheadrightarrow$ is finite,
$K_{T\twoheadrightarrow}$ and $K_{P\twoheadrightarrow}$ must be
finite. Then by Theorem~3, $K_{T\twoheadrightarrow}$ and
$K_{P\twoheadrightarrow}$ can be uniquely decomposed into finitely
many complete sections of knowledge, say
$S_{T1},S_{T2},\ldots,S_{Tn}$ and $S_{P1},S_{P2},\ldots,S_{Pm}$,
respectively. For each $S_{Ti}$ and each $S_{Pj}$, there exists only
one chapter $k_T\otimes k_P$ such that $k_T\subset S_{Ti}$ and
$k_P\subset S_{Pj}$ for $1\leq i \leq n$ and $1\leq j \leq m$. Under
the assumption above, it is certain that the knowledge space
$K_\twoheadrightarrow$ can be decomposed into such chapters. For
each chapter $k_T\otimes k_P$ of knowledge, there must exist two and
only two complete sections $S_{Ti}$ and $S_{Pj}$ of knowledge such
that $k_T\subset S_{Ti}$ and $k_P\subset S_{Pj}$ where $1\leq i \leq
n$ and $1\leq j \leq m$. Therefore, the decomposition is unique.

\medskip

\noindent \textbf{Remark.} One can also define another binary relation on $K_T$ and $K_P$. For
any two knowledge objects $o_1$ and $o_2$, if $o_2$ is related to $o_1$, then $o_2$ is said to be connected to $o_1$. In this case, the relation has no ``direction'' and one can establish similar results for the relation.

\medskip
One of the biggest
challenges for implied knowledge discovery is how to effectively
synthesize procedural knowledge objects from declarative ones. There
are a few studies focused on specific issues, such as derivation of
simple programs or algorithms from given specifications in
particular declarative forms~\cite{Manna,Tudor}, yet developing a
general method or framework to mechanize and automate the process of
synthesis is certainly hard.

\section{Concluding Remarks}

We have introduced the three spaces of data, information, and
knowledge with structures, from which one may observe how domain
information and knowledge can be acquired from data. Some properties
and characteristics of the three spaces are presented and their
interrelations are clarified. More properties about the spaces of knowledge and their subspaces will be investigated further. Our study results in a general
approach for the discovery of knowledge implied in data.

The data space $D$ may be simple and small, while the knowledge
space $K$ built up from $D$ for the given domain $\Delta$ can become
very rich because application of meta-knowledge (mainly the
induction and deduction principles) and domain knowledge to the
information space $I$ may yield many new and valuable knowledge
objects which are related to or may be induced or deduced from $D$
under the interpretation in $\Delta$. Therefore, profound knowledge
can be discovered from $D$ by constructing knowledge objects of $K$
according to specially designed domain-dependent procedures. The
feasibility and effectiveness of our general approach has been
demonstrated by our implementation in the case of geometry
\cite{csw14}, where nontrivial geometric theorems can be discovered
automatically and efficiently from images of diagrams.

To apply our approach to discover knowledge of a concrete domain,
one has to work on several issues, including formalization and
representation of domain knowledge (see, e.g., \cite{cw12}), design
and implementation of induction and deduction schemes, and
interpretation of data and retrieval of information from data in the
domain. We shall report on the results of our studies in selected
domains.


\begin{thebibliography}{33}

\bibitem{sm} Wilson E B. An introduction to scientific research. New York: McGraw-Hill, 1952.

\bibitem{bigdata} Manyika J, Chui M, Brown B, Bughin J, Dobbs R, Roxburgh C, Byers A. Big data: The next frontier for innovation, competition, and productivity. McKinsey Global Institute, 2011.

\bibitem{DMtop10} Wu X D, Kumar V, Quinlan J R, Ghosh J, Yang Q, Motoda H, McLachlan G J, Ng A, Liu B, Yu P S, Zhou Z H, Steinbach M, Hand D J, Steinberg D. Top 10 algorithms in data mining. Knowledge and Information Systems, 2008, 14(1): 1--37

\bibitem{Zhou12} Zhou Z H. Ensemble methods: Foundations and algorithms. Chapman and Hall/CRC, 2012.

\bibitem{sn1} Kleinberg J. Navigation in a small world. Nature, 2000, 406: 845

\bibitem{sn2} Leskovec J, Kleinberg J, Faloutsos C. Graphs over time: densification laws, shrinking diameters and possible explanations. In:
Grossman R, Bayardo R J, Bennett K P, eds. Proceedings of the 11th ACM SIGKDD international conference on Knowledge discovery in data mining, Chicago, USA, 2005. 177--187

\bibitem{Han} Han J W, Pei J, Yin Y W. Mining frequent patterns without candidate generation. Newsletter ACM SIGMOD Record, 2000, 29(2): 1--12

\bibitem{DMsurvey} Liao S H, Chu P H, Hsiao P Y. Data mining techniques and applications -- A decade review from 2000 to 2011. Expert Systems with Applications, 2012, 39: 11303--11311

\bibitem{induction} Russell B. The problems of philosophy. Wilder Publications, 2009.

\bibitem{induction1} Hawthorne J. Inductive logic. In: Zalta E N, ed. The Stanford Encyclopedia of Philosophy, 2012.\\ http://plato.stanford.edu/archives/win2012/entries/logic-inductive/

\bibitem{Karl} Popper K. The logic of scientific discovery (Routledge Classics). Routledge, 2002.

\bibitem{Li} Li W. Mathematical logic: Foundations for information science. Birkh\"{a}user, 2010.

\bibitem{csw14} Chen X Y, Song D, Wang D M. Automated generation of geometric theorems from images of diagrams. Geometric Reasoning --- Special issue of the Annals of Mathematics and Artificial Intelligence. Springer, 2014.

\bibitem{topology} Mendelson B. Introduction to topology: third edition. Dover Publications, 1990.

\bibitem{Carl} Carlsson G. Topology and data. Bulletin of the American Mathematical Society, 2009, 46(2): 255--308

\bibitem{deduction} Sion A. The principle of deduction. TheLogician.net, 2012.\\ http://www.thelogician.net/5\_other\_writings/5\_deduction\_principle.htm

\bibitem{Manna} Manna Z, Waldinger R. A Deductive Approach to Program Synthesis. ACM Transactions on Programming Languages and Systems, 1980, 2: 90--121

\bibitem{Tudor} Dramnesc I, Jebelean T. Automated synthesis of some algorithms on finite sets. In: Proceedings of the 14th International Symposium on Symbolic and Numeric Algorithms for Scientific Computing, Timisoara, Romania, 2012. 143--151

\bibitem{cw12} Chen X Y, Wang D M. Management of geometric knowledge in textbooks. Data \& Knowledge Engineering, 2012, 73: 43--57

\end{thebibliography}
\end{document}